\documentclass[a4paper,10pt,twocolumn]{article}
\usepackage{times}
\usepackage{url}
\usepackage{graphicx}
\usepackage{amsmath}
\usepackage{multirow}
\makeatletter
\def\section{\@startsection {section}{1}{\z@}{-3.5ex \@plus -1ex \@minus -.2ex}{2.3ex \@plus.2ex}{\normalsize\bf\centering}}
\makeatother 
\makeatletter
\def\@seccntformat#1{\csname the#1\endcsname.~}
\makeatother

\makeatletter
\def\subsection{\@startsection {subsection}{2}{\z@}{-3.25ex\@plus -1ex \@minus -.2ex}{1.5ex \@plus .2ex}{\normalsize\bf}}
\makeatother 

\makeatletter
\def\subsubsection{\@startsection {subsubsection}{3}{\z@}{-3.25ex\@plus -1ex \@minus -.2ex}{1.5ex \@plus .2ex}{\normalsize\bf}}
\makeatother 

\title{\Large \bf {
Improving Semi-Supervised Contrastive Learning via Entropy-Weighted Confidence Integration of Anchor-Positive Pairs
}}
\author{
Shogo Nakayama\\
Doshisha University, Kyoto, Japan
\and
Masahiro Okuda\\
Doshisha University, Kyoto, Japan}
\date{\vspace{-9mm}} 

\begin{document}
\maketitle
\thispagestyle{empty} 

\section*{Abstract}

Conventional semi-supervised contrastive learning methods assign pseudo-labels only to samples whose highest predicted class probability exceeds a predefined threshold, and then perform supervised contrastive learning using those selected samples.
In this study, we propose a novel loss function that estimates the confidence of each sample based on the entropy of its predicted probability distribution and applies confidence-based adaptive weighting.
This approach enables pseudo-label assignment even to samples that were previously excluded from training and facilitates contrastive learning that accounts for the confidence of both anchor and positive samples in a more principled manner.
Experimental results demonstrate that the proposed method improves classification accuracy and achieves more stable learning performance even under low-label conditions.

\vspace{1mm}

{\noindent \bf Keywords:} Semi-supervised learning, Contrastive learning, Classification, Entropy weighting
\vspace{6mm}

\section{Introduction}
In recent years, the advancement of deep learning has enabled high-accuracy results in image classification tasks when sufficient labeled data are available \cite{bib:alex}\cite{bib:vgg}\cite{bib:resnet}. However, many challenges remain in situations where labeled data are limited. While unsupervised learning methods have been proposed to utilize unlabeled data \cite{bib:simCLR}\cite{bib:byol}, real-world scenarios typically involve a small amount of labeled data coexisting with a large amount of unlabeled data. Therefore, this study focuses on semi-supervised learning (SSL), which aims to learn effectively from limited labeled data.SSL has great potential to significantly reduce the cost of data annotation, and it has recently become an active research topic \cite{bib:mix}\cite{bib:fix}.
In our previous work \cite{bib:mine}, we achieved performance improvement by effectively combining an MMD(Maximum Mean Discrepancy)-based regularization term with the baseline loss function of \cite{bib:semisup}.
In contrast, the present study aims to enhance performance by directly modifying the original loss function of \cite{bib:semisup}, rather than merely adding a regularization term.

In SSL, a common approach is to assign pseudo-labels to unlabeled data for training. Many studies have also explored extensions and improvements of representative methods such as \cite{bib:fix}. In \cite{bib:fix}, a pseudo-label is assigned to an unlabeled sample when its highest predicted class probability exceeds a predefined threshold. However, if the probability does not surpass the threshold, the sample is discarded and not used for training. To address this issue, semi-supervised contrastive learning methods \cite{bib:semisup} that combine \cite{bib:fix} with supervised contrastive learning \cite{bib:supcon} have been proposed.In \cite{bib:semisup}, a small weight is assigned to samples that cannot be assigned pseudo-labels, and unsupervised contrastive learning is performed using two augmented views derived from the same original image.

However, some samples whose predicted class probabilities do not exceed the threshold may still have reasonably high confidence for certain classes. Excluding these samples entirely could lead to a loss of potentially useful information.
To address this issue, we estimate the confidence of each sample using the entropy of its predicted probability distribution and assign a weight according to this confidence.
Furthermore, we extend the loss function of \cite{bib:semisup} so that the confidence of both anchor and positive samples can be taken into account.
This enables adaptive weighting based on confidence and allows pseudo-label assignment to a wider range of unlabeled samples, leading to more effective utilization of unlabeled data.

\section{Conventional Method}

We first describe the conventional method, namely semi-supervised contrastive learning \cite{bib:semisup}, which we use as the baseline for our experiments. We introduce the notations used throughout this paper and explain how they relate to the existing literature.
\begin{equation}
\scalebox{1}{$
\mathbf{X} =
\begin{bmatrix}
\mathbf{x}^1 
\cdots 
\mathbf{x}^B
\end{bmatrix}
,
\mathbf{y}_x =
\begin{bmatrix}
y_x^1 
\cdots 
y_x^B
\end{bmatrix}
,
\mathbf{U} =
\begin{bmatrix}
\mathbf{u}^1 
\cdots 
\mathbf{u}^{\mu B}
\end{bmatrix}
$}
\end{equation}

Here, $\mathbf{X}$ denotes a mini-batch of labeled samples, and $B$ represents the batch size of the labeled data.
The variable $\mathbf{y}_x$ indicates the class label assigned to each labeled sample.
Similarly, $\mathbf{U}$ denotes a mini-batch of unlabeled samples, and $\mu$ denotes the ratio between the unlabeled and labeled batch sizes.
\begin{equation}
\renewcommand{\arraystretch}{1.5} 
\begin{aligned}
\mathbf{Z}_x &= f(\mathbf{X}) =
\begin{bmatrix}
    \mathbf{z}_x^1, \cdots, \mathbf{z}_x^B
\end{bmatrix}^\top \\[4pt]
\mathbf{Z}_u &=
\begin{bmatrix}
    \mathbf{Z}_{s1} \\
    \mathbf{Z}_{s2}
\end{bmatrix}
= 
\begin{bmatrix}
    f(A_1(\mathbf{U})) \\
    f(A_2(\mathbf{U}))
\end{bmatrix} \\[4pt]
\mathbf{Z}_w &= f(\alpha(\mathbf{U})) =
\begin{bmatrix}
    \mathbf{z}_w^1, \cdots, \mathbf{z}_w^{\mu B}
\end{bmatrix}
\end{aligned}
\renewcommand{\arraystretch}{1.0} 
\label{eq:Z_definitions}
\end{equation}
We denote by $f$ the encoder that maps samples into a hidden space. $A_1$ and $A_2$ denote two independent strong augmentations, while $\alpha$ represents a weak augmentation. $\mathbf{Z}_x$ represents the feature embeddings of labeled data. $\mathbf{Z}_u$ is the set of embeddings obtained by applying two strong augmentations to the unlabeled data. $\mathbf{Z}_w$ represents the set of feature embeddings generated by applying weak augmentations to unlabeled data.
\begin{equation}
\scalebox{1}{$
\mathbf{Z}_c =
\begin{bmatrix}
    \mathbf{z}_c^1, \cdots, \mathbf{z}_c^k
\end{bmatrix},
\quad
\mathbf{y}_c =
\begin{bmatrix}
    y_c^1, \cdots, y_c^k
\end{bmatrix}.
$}
\end{equation}
$\mathbf{Z}_c$ is the set of prototype vectors for each class $k$, where each prototype represents the representative feature of that class in the embedding space. $\mathbf{y}_c$ represents class labels for prototypes.

Next, we introduce the pseudo-label assignment procedure for unlabeled samples, in which class probabilities are derived from cosine similarity, as shown in the following equation.
\begin{equation}
\scalebox{1}{$
p(\mathbf{z}_w^i) := \text{softmax}(\frac{\mathbf{Z}_c \mathbf{z}_w^i} {T'})
$}
\end{equation}
$T'$ is the temperature used during pseudo-labeling. The classification probability for each class is obtained by computing the cosine similarity between the class prototype and the weakly augmented representation of the unlabeled sample.
If the highest probability exceeds a confidence threshold $\tau$, the corresponding class is assigned as the pseudo-label.
For unlabeled samples whose maximum probability does not exceed the threshold, a unique label is individually assigned to each instance.
The pseudo-label for each unlabeled sample is formally defined as follows.
\begin{equation}
\begin{aligned}
c = \arg\max_{c} \, p(\mathbf{z}_w^i)\quad\quad\quad\quad
\\
y_{u}^i =
\begin{cases} 
y_{u}^{\uparrow i}=\begin{bmatrix}
    c,c
\end{bmatrix} & \text{if } \max p(\mathbf{z}_{w}^i) > \tau \\
y_{u}^{\downarrow i} + K & \text{otherwise}
\end{cases}
\end{aligned}
\end{equation}
$c$ represents the class with the highest predicted probability in the distribution.
$K$ is defined as the total number of classes, such as $K$ = 100 in the case of the CIFAR-100 dataset.

Finally, we describe the loss function in detail.
\begin{align}
\label{eq:ssc}
&L_{\mathrm{SSC}}(\mathbf{Z},\mathbf{y},\boldsymbol{\lambda})
= \\
&\frac{1}{\sum_{k\in\mathcal{I}}{\lambda}_k}
\sum_{i\in\mathcal{I}}
\frac{-\lambda_i}{|P(i)|}
\sum_{p\in P(i)} \notag
\log
\frac{
\exp\!\big((\mathbf{z}^{\,i}\!\cdot\!\mathbf{z}^{\,p})/T\big)
}{
\sum_{j\in\mathcal{I}\setminus\{i\}}
\exp\!\big((\mathbf{z}^{\,i}\!\cdot\!\mathbf{z}^{\,j})/T\big)
}
.
\end{align}
where $\mathbf{Z}$ denotes the concatenation of all feature embeddings used in contrastive learning.
$\mathbf{y}$ represents the labels assigned to all samples within the mini-batch, including both ground-truth and pseudo labels.
we define $\mathcal{I} = \begin{bmatrix}1, \dots, N\end{bmatrix}$ as the set of indices corresponding to all samples in the mini-batch, where $N$ is the mini-batch size.
For each anchor $i \in \mathcal{I}$, We define $P(i)$ as the set of indices corresponding to the positive samples for the $i$-th sample (anchor). Specifically, $P(i)$ includes samples in the mini-batch that belong to the same class as the $i$-th sample but does not include the $i$-th sample itself.
the temperature parameter $T$ controls the smoothness of the similarity distribution.

The weight vector $\boldsymbol{\lambda}$ represents the relative importance assigned to different types of data samples.
The values of these weights are determined based on the characteristics of the data, specifically whether a sample is labeled or unlabeled.
The detailed weighting scheme is defined as follows.
\begin{equation}
\begin{alignedat}{2}
&\lambda_x = 1 \:\text{(labeled)},&\quad &\lambda_{u\uparrow}= 1\:\text{(pseudo labeled)} \\
&\lambda_c = 1\: \text{(prototypes)}, &\quad
&\lambda_{u\downarrow}= 0.2\:\text{(unlabeled)}
\end{alignedat}
\end{equation}
This weighting scheme allows the model to effectively leverage both labeled and unlabeled data during training.

\section{Proposed Method}
Based on the semi-supervised contrastive learning framework described above,
we now present our proposed method, which extends the conventional approach through three key components:
a novel loss function that jointly considers the weights of both anchor and positive samples,
entropy-based sample selection,
and adaptive weighting according to the confidence of the predicted class probabilities.
The following subsections describe each component in detail.

\subsection{Proposed Loss Function}

The loss function used in the proposed method is shown below.

{\footnotesize
\begin{align}
\label{eq:ent}
&L_{\mathrm{SSC\text{-}E}}(\mathbf{Z},\mathbf{y},\boldsymbol{\lambda})
= \\
&\frac{1}{\sum_{k\in\mathcal{I}}\overline{\lambda}_k}
\sum_{i\in\mathcal{I}}
\frac{-1}{|P(i)|}
\sum_{p\in P(i)} \notag
\sqrt{\lambda_i \, \lambda_p}
\log
\frac{
\exp\!\big((\mathbf{z}^{\,i}\!\cdot\!\mathbf{z}^{\,p})/T\big)
}{
\sum_{j\in\mathcal{I}\setminus\{i\}}
\exp\!\big((\mathbf{z}^{\,i}\!\cdot\!\mathbf{z}^{\,j})/T\big)
}
\end{align}
}
The notation used in Equations (6) and (8) is identical. In Equation (6), scaling is performed using the sum of anchor weights $\lambda_i$, whereas in Equation (8), the geometric mean of the anchor and positive pair weights $\sqrt{\lambda_i \lambda_p}$ is used as the sample weight, and scaling is performed using the sum of the anchor-wise averaged weights $\overline{\lambda}_i$. By employing this loss function, it becomes possible to compute a contrastive loss that takes into account not only the reliability of the anchor but also that of the positive pairs.
\begin{table*}[t]
    \centering
    \caption{Comparison of accuracy rates. Higher values are shown in bold. “base” denotes the baseline, and “w.ent” denotes the proposed method. “labels/class” indicates the number of labeled samples per class.}
    \label{tab:accuracy_comparison}
    \begin{tabular}{|c|c|cc|cc|}
        \hline
        \multirow{2}{*}{Dataset} & \multirow{2}{*}{labels/class} &\multicolumn{2}{c|}{\textbf{CIFAR-10}}& \multicolumn{2}{c|}{\textbf{CIFAR-100}} \\
        \cline{3-6}
        & & 4 & 25 & 4 & 25 \\
        \hline
         \multirow{2}{*}{Method} & base &0.9441&\textbf{0.9471}& 0.4513 & 0.6444 \\
         & w.ent & \textbf{0.9459} &0.9455&\textbf{0.4639} & \textbf{0.6497} \\
         \hline
        \end{tabular}
        \end{table*}
\subsection{Entropy-Based Sample Selection}
When the predicted class probability of an unlabeled sample exceeds a predefined threshold, the corresponding class is assigned as a pseudo-label. For unlabeled samples that are not assigned pseudo-labels, sample selection and weighting are performed based on the entropy of the predicted probability distribution. The method for computing the entropy of the predicted probability distribution is shown below.
\begin{align}
\mathrm{H}(p(\mathbf{z}_w^i)) &= -\sum_{c=1}^{C} {p(\mathbf{z}_w^i)}_c \log {p(\mathbf{z}_w^i)}_c\\
\mathrm{H_{max}} &= \log C \\
\mathrm{H_{base}} &= \tau_{ent} \cdot \mathrm{H_{max}}
\end{align} 
 $\mathrm{H(p(\mathbf{z}_w^i))}$ denotes the entropy of the predicted probability distribution for each unlabeled sample. $C$ represents the total number of classes. $\mathrm{H_{max}}$ is the maximum entropy of the predicted probability distribution, and $\mathrm{H_{base}}$ is a threshold value. If the entropy of an unlabeled sample falls below $\mathrm{H_{base}}$, the sample is included as a positive pair with labeled or pseudo-labeled samples. Samples whose entropy does not fall below the threshold are treated in the same manner as in the conventional method \cite{bib:semisup}, where they are assigned unique labels and small weights for contrastive learning.
\subsection{Entropy-Based Adaptive Weighting}
\begin{equation}
\lambda_i =
\begin{cases}
1, & \text{if } \mathrm{H}(p(\mathbf{z}_w^i)) = h_i \le e_{\min}, \\[6pt]
w_i, & \text{if } i \in \mathcal{M}_{\mathrm{mid}}.
\end{cases}
\end{equation}

\begin{align} 
s_i &= \frac{\mathrm{H_{base}} - h_i}{\mathrm{H_{base}} - e_{\min}}, \quad i \in \mathcal{M}_{\mathrm{mid}}, \\ 
w_i &= w_{\min} + (1 - w_{\min}) \cdot s_i 
\end{align} 
\[ 
\begin{array}{ll} 
e_{\min} & :~ \text{the largest entropy among pseudo-labeled samples} \\ 
\mathcal{M}_{\text{mid}} & :~ \parbox[t]{6.5cm}{\text{the set of samples subject to}\\ \text{entropy-based weighting}} \\ 
w_{\min} & :~ \text{the minimum value of the weight} 
\end{array} 
\]
This section describes the weighting method based on entropy.
First, if $e_{\min}$ is smaller than $h_i$ in Eq. (12), a weight of $1$ is assigned, as in the case of ordinary pseudo-labeled samples.
For the other samples, weighting is performed using Eqs. (13) and (14).
In Eq. (13), $s_i$ is a scaling factor that ensures the weight value becomes $1$ when $h_i = e_{min}$.
Finally, the weighting is applied using Eq. (14) with $s_i$.

\section{Experiments}
In this section, we describe the experimental setup, implementation details, and results that demonstrate the effectiveness of the proposed method.
\subsection{Dataset}
In this experiment, the CIFAR-10 and CIFAR-100 datasets \cite{bib:cifar} were divided into labeled and unlabeled subsets for use in semi-supervised learning.
\subsection{Comparison Models}
As a baseline, we used the conventional Semi-Supervised Contrastive Learning (SSL) method and compared its image classification accuracy with that of the proposed method described in Section 3.

\subsection{Training Settings}
We used momentum SGD as the optimization algorithm.
The model was trained for 256 epochs, with 1024 steps per epoch.
In the proposed method, entropy-based sample selection and weighting were disabled after 200 epochs,
since samples that still exhibit high uncertainty at that point are likely to contain noisy predictions in the later stages of training.

The momentum coefficient was set to 0.9, and the initial learning rate was 0.03.
The batch sizes for the labeled and unlabeled data were 64 and 448, respectively. In Eq. (11), the parameter $\tau_{ent}$ was set to 0.2 and 0.4 for CIFAR-10 when the number of labeled samples per class was 4 and 25, respectively, and to 0.1 and 0.2 for CIFAR-100 under the same conditions.

We employed a cosine learning rate schedule, where the learning rate $\eta_t$ at step $t$ is determined as follows:
\[
\eta_t = \eta_0 \cos(\frac{7\pi t}{16T})
\]
$\eta_0$ denotes the starting learning rate, and 
$T$ represents the total training epochs.
The network architecture used in all experiments was WideResNet-28-2 \cite{bib:wrn}.
We conducted the experiment with 4 and 25 labeled samples per class to evaluate performance under different label-scarcity conditions.
\subsection{Results}
The experimental results are shown in Table 1.
For CIFAR-10, the proposed method outperforms the baseline when the number of labeled samples per class is 4, but performs slightly worse when it is 25.
In contrast, for CIFAR-100, the proposed method outperforms the baseline in both settings.

\section{Conclusion}
In this study, the effectiveness of the proposed method was validated through experiments conducted on the CIFAR-10 and CIFAR-100 datasets.
In particular, the improvement in classification accuracy was more pronounced when using 4 labeled samples per class 
than when using 25 labeled samples per class. 
However, since the experiments were conducted on only two datasets, the generalizability of the method has not yet been demonstrated.
As future work, we plan to conduct experiments on additional datasets and with different random seeds to further verify the robustness and generality of the proposed method.

\end{document}